\documentclass[conference]{IEEEtran}
\IEEEoverridecommandlockouts
\usepackage{cite}
\usepackage{amsmath,amssymb,amsfonts}
\usepackage{algorithmic}
\usepackage{graphicx}
\usepackage{textcomp}
\usepackage{xcolor}
\usepackage{float}
\usepackage{tabu}
\usepackage{tabularx,booktabs}
\def\BibTeX{{\rm B\kern-.05em{\sc i\kern-.025em b}\kern-.08em
    T\kern-.1667em\lower.7ex\hbox{E}\kern-.125emX}}
\begin{document}

\title{Community Detection Clustering via Gumbel Softmax}

\author{\IEEEauthorblockN{Deepak Bhaskar Acharya}
\IEEEauthorblockA{\textit{Computer Science} \\
\textit{University of Alabama in Huntsville}\\
Huntsville, AL, USA \\
da0023@uah.edu}
\and
\IEEEauthorblockN{Huaming Zhang}
\IEEEauthorblockA{\textit{Computer Science} \\
\textit{University of Alabama in Huntsville}\\
Huntsville, AL, USA \\
zhangh2@uah.edu}
}

\maketitle

\begin{abstract}
Recently, in many systems such as speech recognition and visual processing, deep learning has been widely implemented. In this research, we are exploring the possibility of using deep learning in community detection among the graph datasets. Graphs have gained growing traction in different fields, including social networks, information graphs, the recommender system, and also life sciences. In this paper, we propose a method of community detection clustering the nodes of various graph datasets. We cluster different category datasets that belong to Affiliation networks, Animal networks, Human contact networks, Human social networks, Miscellaneous networks. The deep learning role in modeling the interaction between nodes in a network allows a revolution in the field of science relevant to graph network analysis. In this paper, we extend the gumbel softmax approach to graph network clustering. The experimental findings on specific graph datasets reveal that the new approach outperforms traditional clustering significantly, which strongly shows the efficacy of deep learning in graph community detection clustering. We do a series of experiments on our graph clustering algorithm, using various datasets: Zachary karate club, Highland Tribe, Train bombing, American Revolution, Dolphins, Zebra, Windsurfers, Les Misérables, Political books.

\end{abstract}

\section{Introduction}
\label{intro}
Deep learning has become a hot topic in machine learning and artificial intelligence fields. In specific tasks such as speech recognition, natural language processing, and image processing several algorithms, hypotheses, and large-scale training frameworks for deep learning have been developed and widely implemented. Nevertheless, the use of deep learning in community detection clustering has not yet been thoroughly studied, to our knowledge. The goal of this research is to carry out some preliminary research along that path.

Graphs are structures constructed by a set of nodes(also called vertices) and a set of edges that are relationships between pairs of nodes or vertices. Graph clustering is the process of grouping the nodes of the graph into clusters, taking into account the edge structure of the graph in such a way that there are several edges within each cluster and very few between clusters. Graph clustering intends to partition the nodes in the graph into disjoint groups. Graph clustering has been a long-standing subject of research. 

One of the essential features of real-system graphs is the Community structure, where we consider the graph network communities to be groups of graph nodes with a higher probability of contact. Depending on how well the nodes of the graph are connected, we detect the community structure in the graph network, and group identification has significant consequences for revealing the structure of human social networks or human contact networks. A large number of methods are suggested to solve community structure problem \cite{Wenzhong}. In this article, we have tried a new approach that outperforms the performance of the existing community detection algorithms.

Early methods used various shallow approaches to graph clustering. \cite{girvan2002} used centrality indexes to define community divisions and social communities \cite{social}. \cite{hastings2006community} applied the distribution of opinion to community detection and identified the possible group structure. In the context of this article, graph clustering is nothing but community structure detection and clustering, which is the topic of this paper and not to be mistaken with the clustering of graph sets based on structural similarity.

Very recently, Deepak and Huaming \cite{bhaskar2019feature} proposed feature selection and extraction for Graph Neural Networks, where they are selecting the features which dominate the node classification. Feature extraction and selection for Graph Neural Networks uses a differentiable relaxation of the concrete distribution, and the reparameterization trick \cite{kingma2013auto}. Applying their method along with gumbel softmax approach they selected 225 features out of 1433 features for the Cora citation dataset. They were able to classify with good accuracy using selected 225 features out of 1433 features by using a 2-layered graph convolution network.   

In this paper, we extend the Gumbel softmax approach that \cite{bhaskar2019feature} uses to select and extract the features for the Graph neural network to detect community structure in the graph datasets using deep learning based approach. We conduct a series of experiments using a variety of datasets such as Zachary karate club, Highland Tribe, Train bombing, American Revolution, Dolphins, Zebra, Windsurfers, Les Misérables, Political Books. The result seems to be very impressive with good modularity measure, which is one of the measures of the structure of graphs that we use to measure the strength of the communities or clusters.

We organize the remaining of the paper as follows. In Section \ref{Related work}, we introduce the background and related works in more detail. In Section \ref{Proposed Method}, we introduce our main method. In Section \ref{Experiment Results}, we present the experimental results. In section \ref{conc}, we conclude our work.

\section{Background and related work}
\label{Related work}

\subsection{Community Detection Algorithms}
Community detection is widely researched, and so far, various algorithms have been proposed. Community detection is a method for identifying similar groups and can be a complicated process based on the network nature and scale. Scientists have categorized community detection algorithms in many ways, based on the scale of their research \cite{JAVED201887} has researched the community detection algorithms thoroughly by accepting most of the methods addressed in other surveys.
The algorithm proposed in \cite{Newman_2004} starts with all nodes as individual communities and merge them iteratively to optimize the function 'modularity' to define communities that have many numbers of edges within them and few between them. Many other algorithms in community detection literature, including those suggested by \cite{De_Meo_2011}, heavily focus on optimizing modularity. Label propagation is another well-known identity identification strategy that identifies communities by spreading the labels iteratively across the network. Raghavan, Albert, and Kumara \cite{Raghavan_2007} proposed an algorithm where each node selects a label that has the highest frequency in its 1-neighborhood. These labels are permitted to propagate synchronously and asynchronously around the network until near-equilibrium within the network is reached.
In addition, to assign vertices to clusters, Raghavan employs a local technique based on the majority law. The approach defined in \cite{pons2005computing} uses random walks to identify communities: random walkers usually prefer to remain more within the same group. By using the probability flow of random walks, Rosvall and Bergstrom \cite{Rosvall_2009} approach the problem using an information theoretical perspective to discover communities. Finally, Newman's proposed an approach  \cite{newman2006finding} which is a spectral method based on the modularity matrix's to optimize the modularity metric measure.

\subsection{Gumbel Softmax Approach on Feature Selection}
Deepak and Huaming \cite{bhaskar2019feature} selected Graph Neural Network(GNN) features in the paper feature collection and extraction for Graph Neural Networks, with the citation network datasets. (1) They apply the feature selection algorithm to GNNs using Gumbel Softmax and conduct a series of tests using various comparison datasets: Cora, Pubmed, and Citeseer. (2) They develop a method for ranking the features picked. To show the usefulness of algorithms for both feature selection and ranking of features. Deepak and Huaming demo is an illustrative example for the Cora dataset, where they pick 225 features out of 1433 features and rank them according to prominent features. The proposed deep learning model works well with reduced features, which is around 80-85\ percent decrease in the number of features that the dataset had initially. Results of the experiment reveal that the accuracy slowly decreases by using the selected features falling within range 1-50, 51-100, 100-150, and 151-200 for classification. 

Consider the graph of 'n' nodes and 'f' features. Applying the principle of feature selection also takes down the number of features from $f$ to $k$, where $k$ reflects the number of features picked. First, to train the data collection and characteristics, using the selection matrix of the Gumbel function applied (the matrix that has the features chosen while implementing the Gumbel-Softmax algorithm).

In general, let $X_{n \times f}$ be the matrix of the input features where 'n' represents the total number of nodes, and 'f' represents the total number of features in the graph data set for each node. Consider $M_{f \times k}$ where 'f' represents the total number of features in the dataset of the graph, and 'k' denotes the features that we select from 'f' features.

Using the Gumbel Softmax function and the method proposed, Deepak and Huaming select the features in a graph citation dataset. Gumbel-Softmax distribution is \cite{Shixiang}, "a continuous distribution over the simplex which can approximate samples from a categorical distribution". A categorical distribution, by defining the highest probability to one, and all the other probability to zero is a one-hot vector.

Let z be a categorical, random variable with probabilities of the form $\pi_1, \pi_2, \cdots,\pi_k$. Julius \cite{Julius} uses Gumbel-Max trick which provides an easy and efficient way to draw samples z from a categorical distribution with the stated class probabilities $\pi$:

\begin{equation} \label{argmax}
z = one\_hot (arg max_i[g_i + log \pi_i])
\end{equation}

Training a neural network by gradient descent requires the differentiation of each network operation. Remember that in Equation \ref{argmax}, where $i = 1,2,\cdots, k$, the argmax function and the stochastic sampling process $g_i$ are not differentiable. Next, an optimal solution for having argmax differentiable is to approximate it by a softmax function. One can also use a temperature of $\tau$ to control the argmax approximation standard as follows:

\begin{equation} \label{actual_gumbel}
y_i=\frac{exp((log(\pi_i)\ +\ g_i)/\tau)}{\sum_{j=0}^{k}{exp((log(\pi_j)\ +\ g_j)/\tau)}} \ for \ i = 1,2,3, ...... ,k.
\end{equation}

The two layer Graph Convolution Network (GCN) used in the experiment is defined as 
\begin{equation} \label{gcn_1}
GCN(X,A) = Softmax(A (ReLu(AXW_GW_1)) W_2) 
\end{equation}
To verify the selected features and calculate the accuracy for classification they use the following two layer Graph Convolution Network as defined below
\begin{equation} \label{gcn_2}
GCN(X,A) = Softmax(A (ReLu(AXW_G')) W_2) 
\end{equation}
$A$: Adjacency matrix of the undirected graph G.  \newline
$X$: Input feature matrix. \newline
$W_G$: Gumbel-Softmax feature selection / feature extraction matrix. \newline
$W_G'$: feature selection / feature extraction matrix obtain
ed from the result of Equation \ref{gcn_1}.\newline
$W1,W2$: Layer-specific trainable weight matrix.\newline
$ReLu$ : Activation function ReLu(.) = max(0,.).

\section{Proposed Method}\label{Proposed Method}
 Consider the graph of 'n' nodes and the adjacency matrix 'A'. An adjacency matrix is a square matrix used to describe a finite graph. Matrix elements signify whether or not the pairs of vertices in the graph are adjacent. Then, applying our idea, we cluster the graph into k clusters.

The method used in the experiment is defined as below:
\begin{equation} \label{gcn_2}
Graph-Cluster(A) = Softmax(W_C^tAW_C) 
\end{equation}

In the Equation \ref{gcn_2}, '$A$' indicates the Adjacency matrix of the undirected graph G and
'$W_C$' indicates the Gumbel cluster weight matrix.

In general, let $A_{n \times n}$ be the adjacency matrix where 'n' represents the total number of nodes in the graph dataset. The size of the matrix $W_C$ is $n \times k$, where k indicates the number of clusters. When we perform $W_C^tAW_C$ operation, we obtain a matrix of the size $k \times k$ and let us call this matrix as $R_{k \times k}$. The resultant $R_{k \times k}$ matrix shows the strength of the cluster where the primary diagonal shows the strength of data points within cluster groups, and other elements of the matrix $R$ provide details on the strength of data points between different cluster groups. Then, we apply softmax function on the obtained matrix $R$ to express our inputs as a discrete probability distribution. Mathematically, this is defined as follows:
\begin{equation} \label{Softmax}
Softmax(x_i)=\frac{exp(x_i)}{\sum_{j=0}^{m}{exp(x_j)}} \ for \ i = 1,2,3, ...... ,m.
\end{equation}

Now, consider the trained Gumbel cluster weight matrix($W_c$), which is of the size {$n \times k$}. Here, each row is a graph node and will sum up to 1, and the index of the maximum row value is the cluster to which the row node belongs. For example, let us consider k = 2, i.e., we are trying to cluster the dataset into 2 cluster groups. Here, we assume the first row data as [0.9 0.1], where 0.9 is at the $0^{th}$ index, and 0.1 is at the $1^{st}$ index. Looking at the index of the maximum value in the row vector, one can easily say that the graph node belongs to cluster 0. If the second-row assumed data is [0.29 0.71], where 0.29 is at the $0^{th}$ index, and 0.71 is at the $1^{st}$ index. Looking at the index of the maximum value in the row vector, one can easily say that the graph node belongs to cluster 1. Likewise, we look into all the rows and then cluster all the nodes of the graph dataset into the cluster group. The row data values indicate the influence of the graph node towards the cluster group.

The result obtained from the equation \ref{gcn_2} is compared with the loss function. The loss function for our experiment used is the identity matrix($I_{k \times k}$) where each diagonal element represents the cluster group. 
 
Using the gumbel softmax function and the method proposed, we find the community cluster of the graph dataset nodes. 

\section{Experiment Results}\label{Experiment Results}
We cluster different category dataset that belongs to Affiliation networks, Animal networks, Human contact networks, Human social networks, Miscellaneous networks. The datasets are downloaded from Konect \cite{konect}.
    \subsection{Datasets}
    
     \subsubsection{Human Social Networks} Human social networks are real-world social networks among human beings. The links are offline, and not from a social network.\hfill\\
        
        \textbf{Zachary Karate Club(Figure \ref{Karate-1} and Figure \ref{Karate-2}):} The network data were collected from members of the University Karate Club by Wayne Zachary in 1977. Every node represents a member of the club, and an edge represents a link between the two members of the club. The network is undirected. The often-discussed problem of using this dataset is to find the two groups of people to whom the karate club splits after an argument between two teachers.\hfill\\
        
        \textbf{Highland Tribes(Figure \ref{Tribe-1} and Figure \ref{Tribe-2}):} The network is the Gahuku – Gama alliance system of the Eastern Central Highlands of New Guinea, signed social network of tribes from Kenneth Read (1954). The network comprises seventeen tribes connected by friendship ("Rova") and enmity ("Hina").\hfill\\
        
        \textbf{Train Bombing(Figure \ref{TrainBomb-1} and Figure \ref{TrainBomb-2}):} The network is undirected, as retrieved from newspapers, and includes communications between alleged terrorists involved in the Madrid train bombing on 11 March 2004. A node represents a terrorist, and an edge between two terrorists shows that the two terrorists had a connection. The weights on edge indicate how 'strong' relation was. The relationship includes friendship and co-participating in training camps or previous attacks.\hfill\\

    \subsubsection{Affiliation Networks}
    Affiliation networks are the networks denoting the membership of actors in groups.\hfill\\
    
        \textbf{American Revolution(Figure \ref{american_revul-1} and Figure \ref{american_revul-2}):} The network includes membership records of 136 people in 5 organizations from the pre-American Revolution era Figure. The graph contains well-known persons such as US activist Paul Revere. An edge between an individual and an agency suggests the individual was an organization member and is represented in the Figure \ref{american_revul-1}.\hfill\\

    \subsubsection{Animal Networks}Animal networks are networks of animal communications. They are the animal equivalent to human social networks. \hfill\\
    
        \textbf{Dolphins(Figure\ref{Dolphin-1} and Figure \ref{Dolphin-2}):} The network includes a Bottlenose Dolphin Social Network. The nodes are the bottlenose dolphins (genus Tursiops) of a group of bottlenose dolphins live off Doubtful Sound, a New Zealand fjord (spelled fiord in New Zealand). An edge suggests the interaction is regular. The dolphins were observed from 1994 through 2001.\hfill\\
        
        \textbf{Zebra(Figure \ref{Zebra-1} and Figure \ref{Zebra-2}):} The network involves networks of animals from the group that communicates to each other. They are the animal equivalent to human social networks. Note that website datasets such as Dogster are excluded here in the category of Social networks because humans generate the networks.\hfill\\
        
    \subsubsection{Human Contact Networks} Human communication networks are real networks of interaction between people, i.e., talking to each other, spending time together, or at least being physically close. More often than not, by giving out RFID labels to individuals with chips that monitor whether other individuals are nearby, these datasets are collected..\hfill\\
    
    \textbf{Windsurfers(Figure \ref{windsurfers-1} and Figure \ref{windsurfers-2}):} This undirected network includes interpersonal communications during the Fall of 1986 between windsurfers in southern California. A node represents a windsurfer, and an edge between two windsurfers indicates an interpersonal interaction occurred.\hfill\\

    \subsubsection{Miscellaneous Networks} Miscellaneous networks are any networks that do not fit into one of the other categories.\hfill\\
         \textbf{Political Books(Figure \ref{polbooks-1} and Figure \ref{polbooks-2}):} The network is of books published around the time of the 2004 presidential election about US politics and distributed through online bookseller Amazon.com. Edges between books reflect repeated co-purchases by the same purchasers of books.\hfill\\
         
        \textbf{Les Misérables(Figure \ref{Lesmiserables-1} and Figure \ref{Lesmiserables-2}):} This undirected network includes character co-occurrences in the novel 'Les Misérables' by Victor Hugo. A node represents a character, and an edge between two nodes suggests that these two characters existed in the book's same chapter. The weight of each relation reveals how much this kind of co-appearance happened.\hfill\\
    
    \subsection{Results}
    To evaluate the effectiveness of Community Detection Clustering via Gumbel Softmax(CDCGS), we have carried out numerous tests on different types of network benchmarks. The same cluster nodes share the same color. 
    
    The algorithms are getting very different results, particularly for the number of established groups. We compare our proposed method results with the results of algorithms such as greedy optimization of modularity(GOM) \cite{Clauset_2004}, leading eigenvector of the community matrix(ECM) \cite{newman2006finding}, edge betweenness(EB) \cite{PhysRev}, short random walks(RW) \cite{pons2005computing}, Infomap community finding(ICF) \cite{Rosvall_2009}, multi-level optimization of modularity(MOM) \cite{Blondel_2008}, propagating labels(PL)  \cite{Raghavan_2007}. To complete the analysis, we also computed adjusted randomized index(ARI), normalized mutual information(NMI), homogeneity(HOMO), completeness(COMP), and v-measure(V-MES) \cite{rosenberg2007v}. To evaluate the identification of metadata groups by various algorithms, we took the best scores of the Zachary karate club network along with different algorithms. Our method CDCGS outperforms ECM, EB, RW, MOM, and PL algorithms with the metric results, and we have presented them in Table \ref{Table_result2}.
    
    We do not know the specific communities with any of the other networks, or we cannot use the metrics as they are not effective. Instead, we use the modularity metric \cite{PhysRev}, which does not require the actual group assignment and is based only on network structure. The Modularity measure uses the fraction of edges that link vertices within the same group.  Table \ref{Table_result} displays tests result for all algorithms, including CDCGS, which is proposed by us on all of the networks. CDCGS outperforms algorithm which performs best on Zachary karate, Dolphin, Highland Tribes, and Windsurfers networks, have given the best modularity. CDCGS performs almost near to the highest modularity achieved with the other dataset networks. 
    
\begin{figure}[h]
\centering
    \includegraphics[scale=0.5]{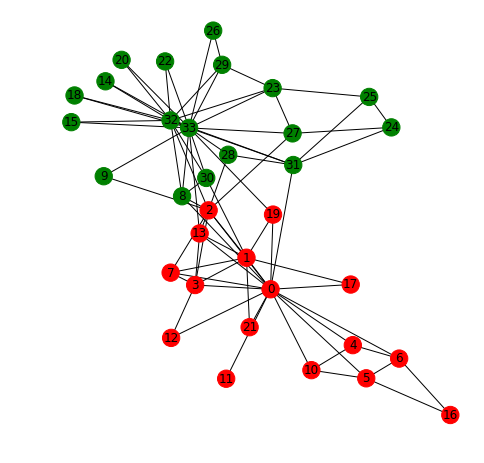}
    \caption{Zachary’s Karate Club Network with 2-Clusters} 
    \label{Karate-4}
\end{figure}

    The ground truth community structure of the Zachary karate club has 2-clusters. When we cluster Zachary network into 2-clusters(Figure \ref{Karate-4}) then metric values ARI = 1, NMI = 1, HOMO = 1, COMP = 1, V-MES = 1 and the modularity measure = 0.371. Still, We identified 4-clusters that give us the modularity measure of 0.4197. Higher modularity values imply a better assignment to the cluster.
    
     \begin{table*}
     \caption{Zachary karate club network Metrics for the different algorithms and our method(CDCGS) }
        \label{Table_result2}
        \begin{tabu} to \textwidth {XXXXXXXXX}
            \toprule
           \textsc{\textbf{Metric}} & 
            \textsc{\textbf{GOM}} & 
            \textsc{\textbf{ECM}}&
            \textsc{\textbf{EB}}&
            \textsc{\textbf{RW}} & 
            \textsc{\textbf{ICF}} & 
            \textsc{\textbf{MOM}} &
            \textsc{\textbf{PL}} &
            \textsc{\textbf{CDCGS}}\\
            \midrule
            ARI  & 0.6802  & 0.5120 & 0.5125      & 0.2620 &0.7021 &0.4619 &0.4714 &0.5414\\
             NMI  & 0.6924  & 0.6770 & 0.6097      & 0.3729 &0.6994 &0.5866 &0.5282 &0.6873\\ 
              HOMO  & 0.8664  & 1 & 0.8850       & 0.5588 &0.8535 &0.8535 &0.6905 &1\\ 
              COMP & 0.5766  & 0.5118 & 0.4651     & 0.2798 &0.5925 &0.44.68 &0.4277 &0.5235\\ 
              V-MES & 0.6924  & 0.6770 & 0.6097     & 0.3729 &0.6994 &0.5866 &0.5282 &0.6872\\ 
            \bottomrule
        \end{tabu}
    \end{table*}
    
    \begin{table*}
     \caption{BEST MODULARITY RESULTS OBTAINED BY OUR METHOD(CDCGS), AND OTHER ALGORITHMS FOR THE REAL WORLD NETWORKS}
        \label{Table_result}
        \begin{tabu} to \textwidth {X[3] XXXXXXXX}
            \toprule
            \textsc{\textbf{Dataset}} & 
            \textsc{\textbf{GOM}} & 
            \textsc{\textbf{ECM}}&
            \textsc{\textbf{EB}}&
            \textsc{\textbf{RW}} & 
            \textsc{\textbf{ICF}} & 
            \textsc{\textbf{MOM}} &
            \textsc{\textbf{PL}} &
            \textsc{\textbf{CDCGS}}\\
            \midrule
            ZACHARY KARATE  & 0.3806  & 0.3934 & 0.4012 & 0.3532 &0.4020 &0.4188 &0.3990 &\textbf{0.4197}\\ 
            DOLPHIN  & 0.4954  & 0.4911 & 0.5193 & 0.4888 &0.5249 &0.5185 &0.4512 &\textbf{0.5246}\\ 
            LES MISÉRABLES  & 0.5000 & 0.5322  & 0.5380 & 0.5214 & 0.5461 &\textbf{0.5555} &0.5367 &0.5389\\ 
             TRAIN BOMBING  & 0.4102  & 0.4055 & 0.3955 & 0.4146 &0.4402 &\textbf{0.4483} &0.4295 &0.4376\\
            HIGHLAND TRIBES  & 0.1603  & 0.1527 & 0.085 & 0.1052 &0 &0.1529 &0 &\textbf{0.1612}\\
           AMERICAN REVOLUTION &\textbf{0.5820} &0.5722 &0.5754 &0.5731 &0.5817 &0.5820 &0.5170 &0.5812\\
           ZEBRA &\textbf{0.2768} &0.2768 &0.2602 &0.2702 &0.2606 &0.2768 &0.2606 &0.2702\\
           WINDSURFERS &0.2581 &0.2545 &0.2302 &0.2581 &0.0 &0.2545 &0.2581 &\textbf{0.2581}\\
           POLITICAL BOOKS &0.5019 &0.4671 &0.5069 &0.5228 &\textbf{0.5228} &0.5204 &0.4873 &0.5208\\
            \bottomrule
        \end{tabu}
    \end{table*}

\begin{figure}[h]
\centering
    \includegraphics[scale=0.5]{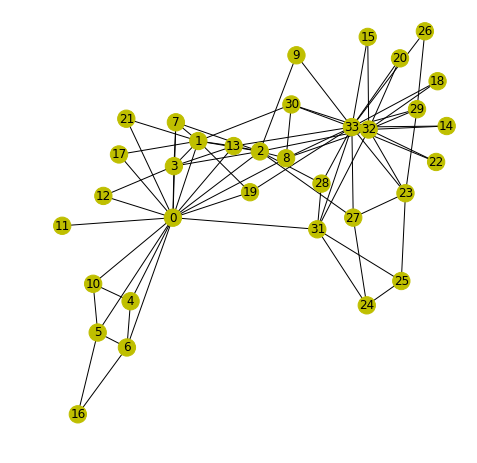}
    \caption{Zachary’s Karate Club Network} 
    \label{Karate-1}

\end{figure}

\begin{figure}[h]
\centering
    \includegraphics[scale=0.5]{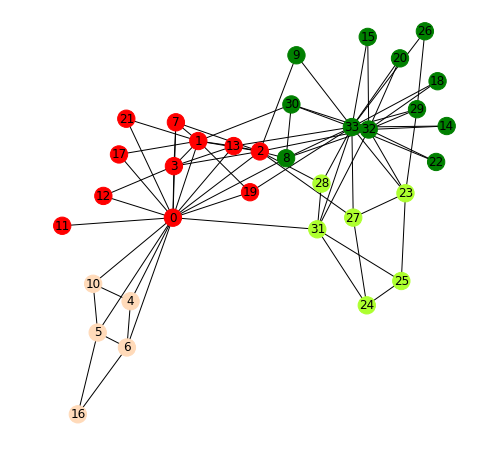}
    \caption{Zachary’s Karate Club Network with 4-Clusters} 
    \label{Karate-2}
\end{figure}

\begin{figure}[h]
\centering
    \includegraphics[scale=0.35]{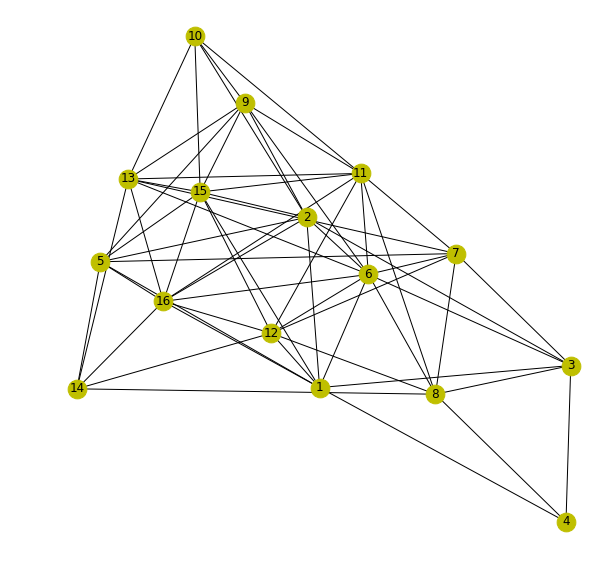}
    \caption{Highland Tribes Network} 
    \label{Tribe-1}
\end{figure}

\begin{figure}[h]
\centering
 \includegraphics[scale=0.35]{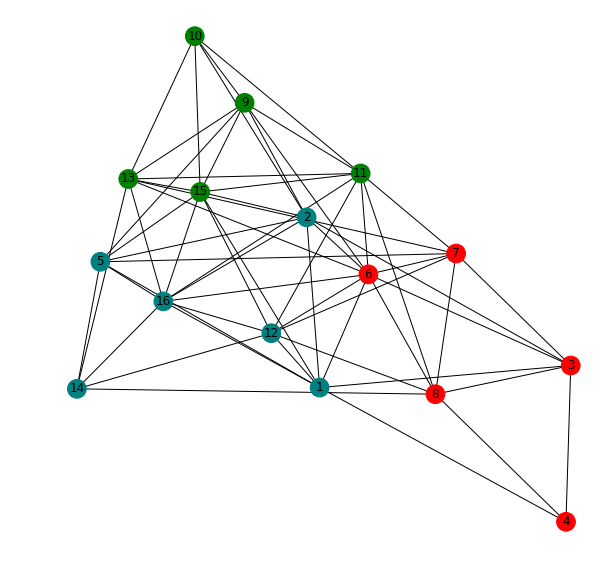}
    \caption{Highland Tribes Network with 3-Clusters} 
    \label{Tribe-2}
\end{figure}

\begin{figure}[h]
\centering
\includegraphics[scale=0.42]{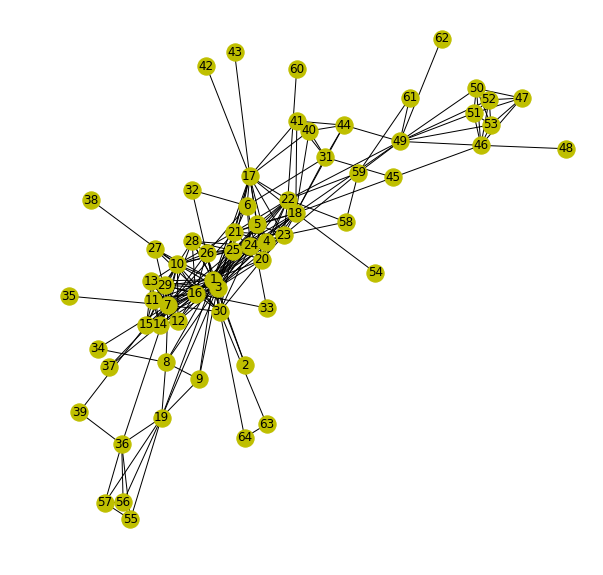}
    \caption{2004 Madrid Commuter Train Bombing Terrorist Network} 
    \label{TrainBomb-1}
\end{figure}

\begin{figure}[h]
\centering
\includegraphics[scale=0.42]{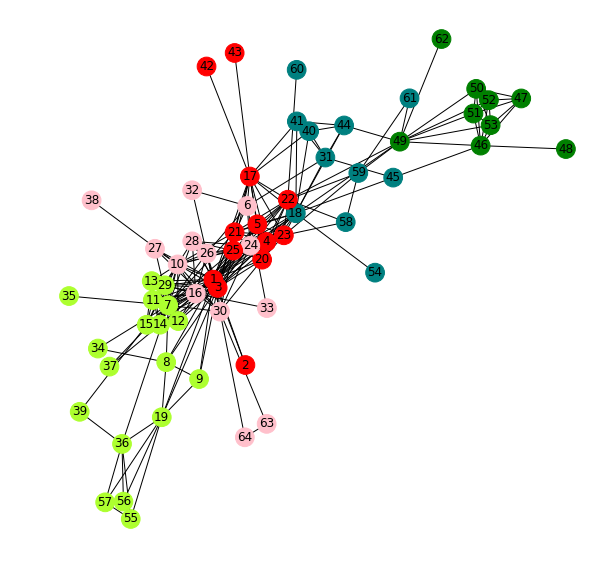}
    \caption{2004 Madrid Commuter Train Bombing Terrorist Network with 4-Clusters} 
    \label{TrainBomb-2}
\end{figure}

\begin{figure}[h]
\centering
  \includegraphics[scale=0.45]{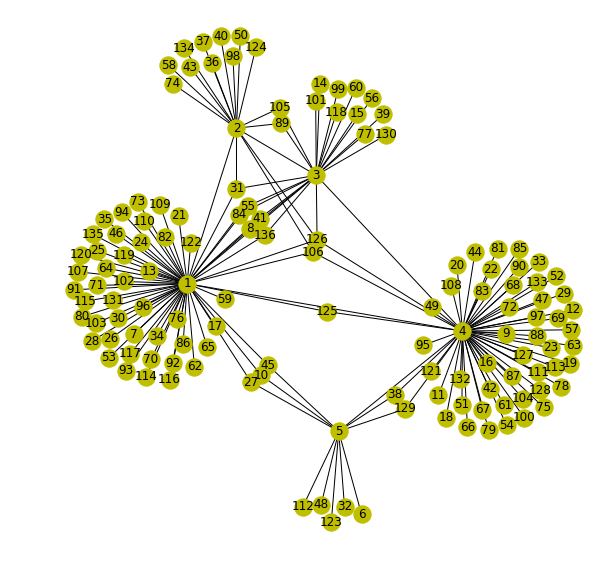}
    \caption{American Revolution Network} 
    \label{american_revul-1}  
\end{figure}

\begin{figure}[h]
\centering
  \includegraphics[scale=0.45]{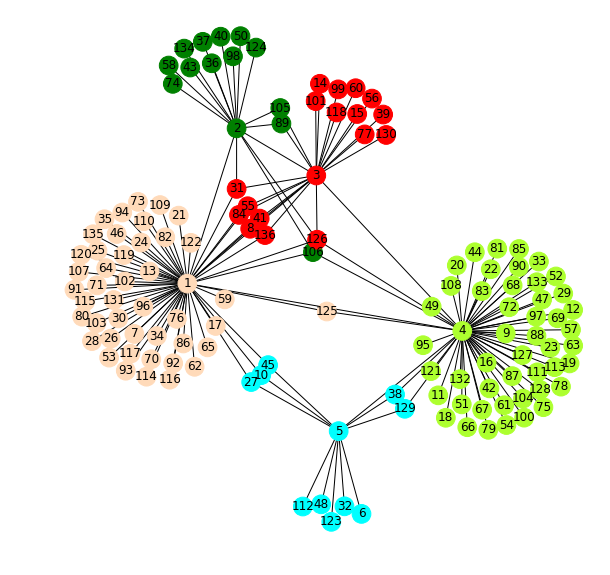}
    \caption{American Revolution Network with 5-Clusters} 
    \label{american_revul-2}  
\end{figure}

\begin{figure}[h]
\centering
 \includegraphics[scale=0.42]{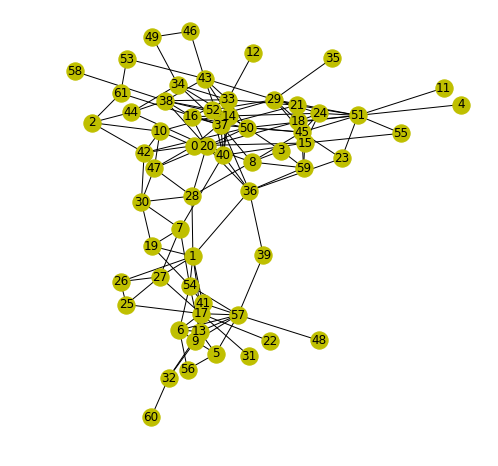}
    \caption{Dolhpin Communication Network} 
    \label{Dolphin-1}
\end{figure}

\begin{figure}[h]
\centering
\centering
 \includegraphics[scale=0.42]{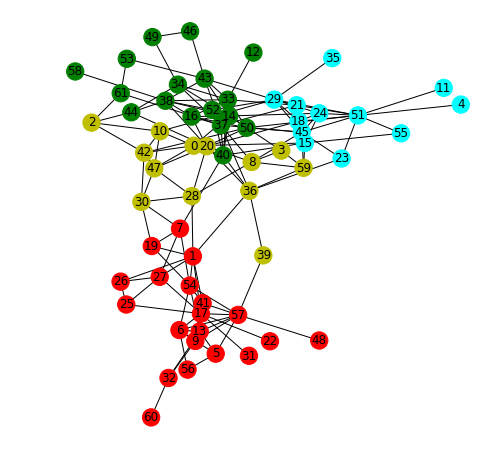}
    \caption{Dolhpin Communication Network with 
    4-Clusters} 
    \label{Dolphin-2}
\end{figure}

\begin{figure}[h]
\centering
  \includegraphics[scale=0.7]{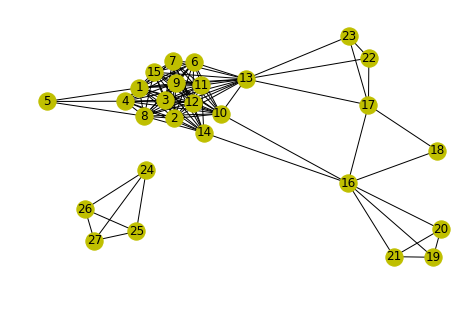}
    \caption{Zebra Communication Network} 
    \label{Zebra-1}  
\end{figure}

\begin{figure}[h]
\centering
  \includegraphics[scale=0.7]{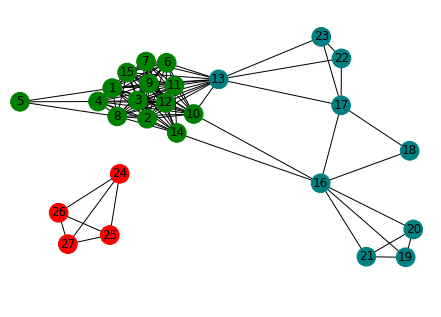}
    \caption{Zebra Communication Network with 3-Clusters} 
    \label{Zebra-2}  
\end{figure}

\begin{figure}[h]
\centering
  \includegraphics[scale=0.35]{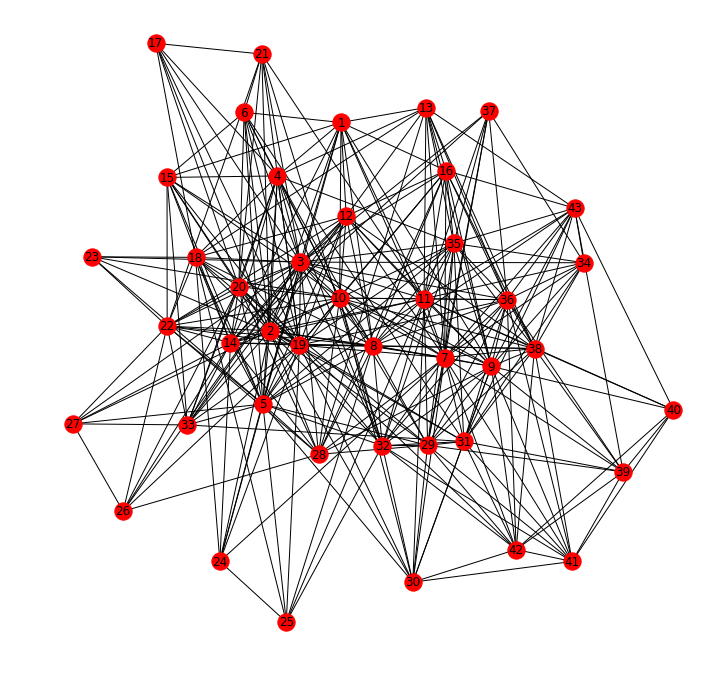}
    \caption{Windsurfers Network} 
    \label{windsurfers-1}  
\end{figure}

\begin{figure}[h]
\centering
  \includegraphics[scale=0.35]{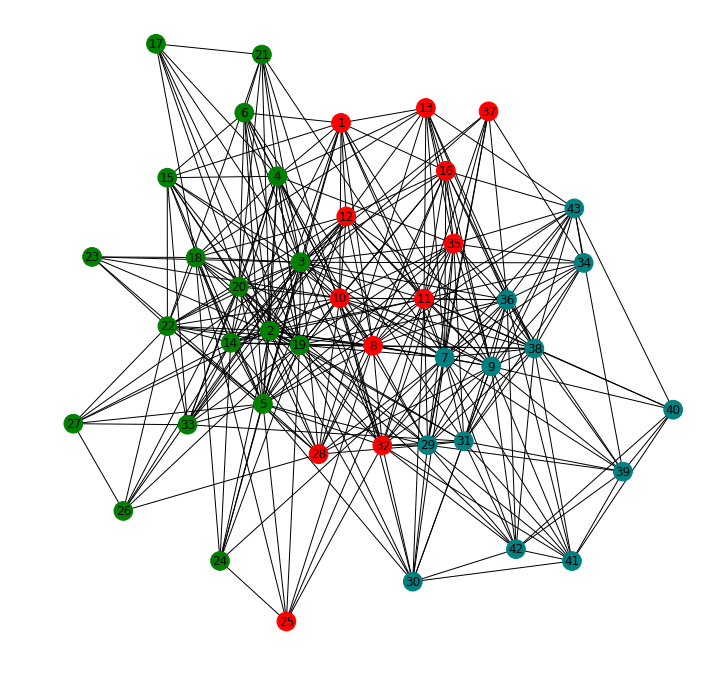}
    \caption{Windsurfers Network with 3-Clusters} 
    \label{windsurfers-2}  
\end{figure}

\begin{figure}[h]
\centering
 \includegraphics[scale=0.4]{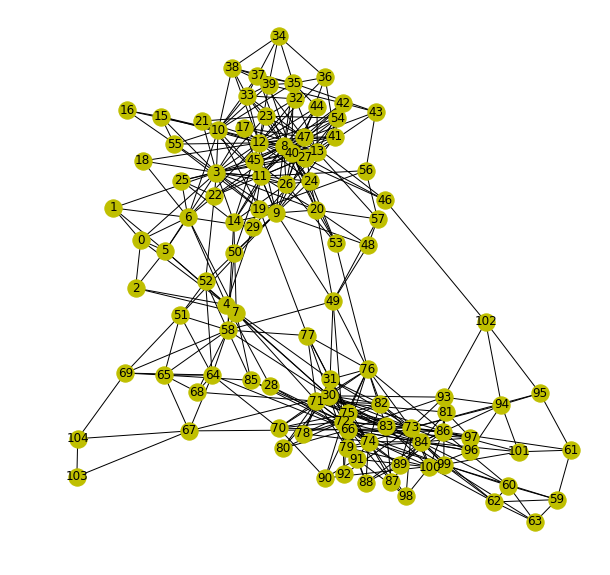}
    \caption{Political Books Network} 
    \label{polbooks-1}
\end{figure}

\begin{figure}[h]
\centering
 \includegraphics[scale=0.4]{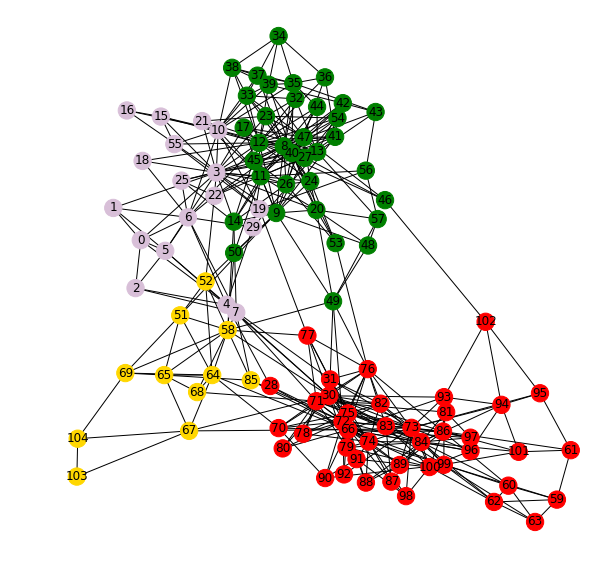}
    \caption{Political Books Network with 4-Clusters} 
    \label{polbooks-2}
\end{figure}

\begin{figure}[h]
\centering
 \includegraphics[scale=0.45]{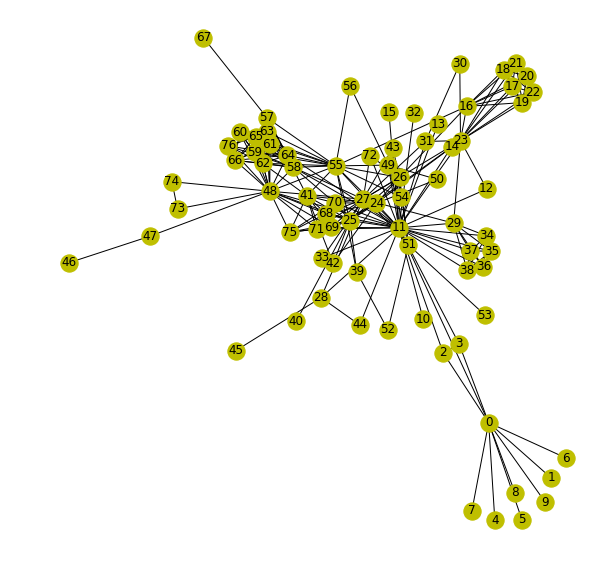}
    \caption{Les Misérables Network} 
    \label{Lesmiserables-1}
\end{figure}

\begin{figure}[h]
\centering
 \includegraphics[scale=0.35]{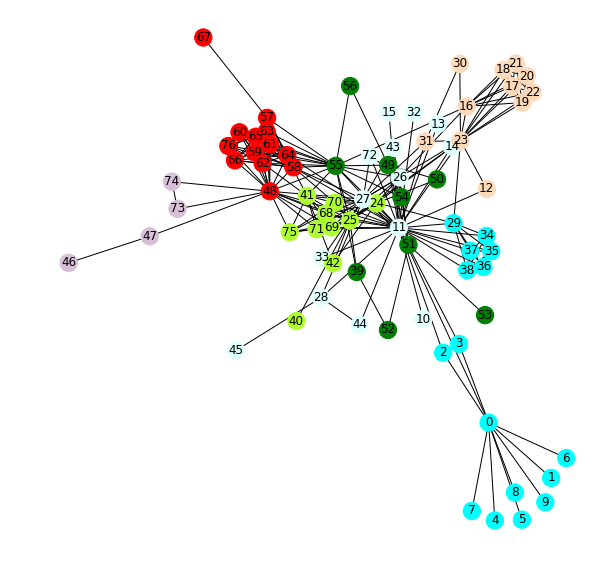}
    \caption{Les Misérables Network with 7-Clusters} 
    \label{Lesmiserables-2}
\end{figure}

     Figures \ref{Karate-1}, \ref{Tribe-1}, \ref{TrainBomb-1}, \ref{american_revul-1}, \ref{Dolphin-1}, \ref{Zebra-1}, \ref{windsurfers-1}, \ref{polbooks-1} and \ref{Lesmiserables-1} shows the dataset graph before clustering. Figures \ref{Karate-2}, \ref{Tribe-2}, \ref{TrainBomb-2}, \ref{american_revul-2}, \ref{Dolphin-2}, \ref{Zebra-2}, \ref{windsurfers-2}, \ref{polbooks-2} and \ref{Lesmiserables-2} shows the beauty of the proposed method after community detection clustering. Here, we can cluster a supervised, unsupervised graph dataset.

\section{Conclusion}
\label{conc}
We introduced the community detection clustering via the gumbel softmax strategy for the different types of Graph network datasets. The experimental findings demonstrate the effectiveness by choosing appropriate parameter values and evaluating the consistency of the resultant clustering. The method currently available is just as diverse as the graph clustering applications. Thus, the suggested algorithm can be regarded as a feasible and effective algorithm for seeking optimal community detection problem solutions. The research completed, however, is on a dataset of unweighted and undirected graphs. We are currently experimenting with applying the principle of clustering for Graphs on a weighted and directed graph dataset.

\bibliographystyle{abbrv}
\bibliography{bib-sam}
\end{document}